# A Novel Correlation-optimized Deep Learning Method for Wind Speed Forecast

Yang Yang[1], Jin Lang[2], Jian Wu[3], Yanyan Zhang[4], Xiang Zhao[1]

1. Key Laboratory of Data Analytics and Optimization for Smart Industry (Northeastern University), Ministry of Education, China, 110819. {yang_cmu@icloud.com; lesliezhaox@gmail.com}
2. Liaoning Engineering Laboratory of Data Analytics and Optimization for Smart Industry, Northeastern University, Shenyang, China, 110819.{langjin@ise.neu.edu.cn}
3. Liaoning Key Laboratory of Manufacturing System and Logistics, Northeastern University, Shenyang, China, 110819.{wujian@mail.neu.edu.cn}
4. Institute of Industrial & Systems Engineering, Northeastern University, Shenyang, China, 110819.{zhangyanyan@ise.neu.edu.cn}

**Abstract**

The increasing installation rate of wind power poses great challenges to the global power system. In order to ensure the reliable operation of the power system, it is necessary to accurately forecast the wind speed and power of the wind turbines. At present, deep learning is progressively applied to the wind speed prediction. Nevertheless, the recent deep learning methods still reflect the embarrassment for practical applications due to model interpretability and hardware limitation. To this end, a novel deep knowledge-based learning method is proposed in this paper. The proposed method hybridizes pre-training method and auto-encoder structure to improve data representation and modeling of the deep knowledge-based learning framework. In order to form knowledge and corresponding absorbers, the original data is preprocessed by an optimization model based on correlation to construct multi-layer networks (knowledge) which are absorbed by sequence to sequence (Seq2Seq) models. Specifically, new cognition and memory units (CMU) are designed to reinforce traditional deep learning framework. Finally, the effectiveness of the proposed method is verified by three wind prediction cases from a wind farm in Liaoning, China. Experimental results show that the proposed method increases the stability and training efficiency compared to the traditional LSTM method and LSTM/GRU-based Seq2Seq method for applications of wind speed forecasting.

**Highlights**

1. A novel deep knowledge-based learning network is proposed and applied in wind speed sequence forecast.
2. In order to develop deep knowledge learning framework, an optimization model based on correlation is proposed, which increases the cognition and memory units of knowledge in the traditional deep learning framework.
3. Objective function on Pearson correlation coefficient is optimized by convex transformation.
4. Performance and reliability are verified by real case studies.

## 1. Introduction

The overconsumption and exhaustion of fossil fuel has led to an increasing demand for clean renewable energy. Wind power is often used as an answer to the demands for clean energy development and is highly anticipated and persistently supported by governments all over the world [1][2]. Nevertheless, as many other clean energy sources, wind power is not a stable energy source. Without accurate prediction, power source cannot be safely fed into the grid as well as some risks for daily maintenances. Thus, in the process of wind power utilization, the main task is to forecast wind speed and wind power over a period of time in the future. However, the wind involves complex spatiotemporal features, which is generally considered as intermittent and random in academic research [3][4][5]. Furthermore, though the wind prediction has been studied for ages from the previous numerical meteorological models and time series statistical models to the recent researches represented by machine learning and hybrid methods, there still exist lots of obstacles to achieve accurate and reliable forecast under practical circumstances. Even now, wind prediction is an urgent and imperative research area with many direct applications such as solar energy, hurricane and forest fire prediction and other extensible spatiotemporal forecasting areas.

There exists an abundance of research into wind prediction methods and their applications that mainly concentrate on the improvement of model structures or the enlargement of data dimensions. Origins of spatio-temporal correlation methods come from Alexiadis *et al.* 's research [4] on artificial neural network models for forecasting wind speed in 1999. Before their research, the mainstream wind prediction methods were basically according to time series sequences of the certain [7]. Recently, the academic focus on forecast models has been turning from the physical and statistical models [6][7] to deep learning models [8][9][10].

As for modern research, some studies on wind speed forecast serve the power markets or the multi-energy scheduling and management systems of power grid. These modern methods are inclined probabilistic models based on the stochastic process. A scenario tree model [11][12] for multistage stochastic programs has been proposed and developed into power management problems with considering German market prices.

Furthermore, for optimal integration of wind energy into power, spatiotemporal analysis has been widely used to improve the probabilistic models. Tastu *et al.* established a methodology with probabilistic wind power forecasts and it improved reliability for forecast uncertainty [13]. He *et al.* put forward Markov chain-based stochastic models based on the statistical distribution of aggregate power with a graph learning-based spatiotemporal analysis [14]. Further in studies on spatial statistics, Wytock and Kolter [15] developed a sparse Gaussian conditional random field for energy forecasting problem with convex learning methods. Wang *et al.* introduced the deep belief network based deterministic and probabilistic approaches for the wind speed prediction [8]. The above methods are established on statistics of probability rather than modeling of probability. In order to solve practical problems, the statistics of probability have to be simplified. Therefore, it is insufficient to extract the additional features from the nonlinear and non-stationary data.

In addition, more and more scientists have discussed the relationship between deep learning models and probabilistic graph models using mathematical theory to support and optimize deep learning model [16]. Deep learning [17] is one of the most effective novel forecasting methods that can process and synthesis the rich historical data. It is shown that deep learning methods are gradually applied in many areas including energy forecasting. Especially, for the tasks or competitions based on fixed training data such as image recognition or semantics analysis, the prediction accuracy has been improved significantly by deep learning model. In 2016, Hu *et al.* introduced transfer learning method [18] for short-term wind speed prediction with deep neural network, which reveals that the learning ability of deep learning model performed better than some shallow models, such as SVR and ELM, while there is a certain dependence on data. Then, there were some other scholars who continued to improve the applications based on advanced deep learning methods in the field of wind prediction. Ghaderi *et al.* proposed a deep learning method based the structure of Recurrent Neural Network (RNN) and Long Short Term Memory (LSTM) [5]. This model took wind speed, direction, environment and neighborhoods into consideration. Manero *et al.* has applied sequence to sequence (Seq2Seq) model to wind power prediction problem [9].

The deep learning model can not only inherit methodological relation but also would like to introduce more practical historical data from the real world so that reduce the gap between model and reality [18]. However, concerning the instability of wind and the complication of actual wind power, they are still not expected to be reliable enough to guarantee practical applications.

Thus, we hope to focus on improving the stability of deep learning methods for wind prediction in this work. A novel knowledge-based deep learning method is proposed in this paper. Based on Pearson correlation coefficient, an optimized correlation-based model has been built to form structured knowledge. Then, the model is optimized by convex transformation and convex optimization methods to obtain simplified and efficient deep networks with structured knowledge in different periods. Finally, the networks are regarded as the encoder layers of a deep learning framework (Seq2Seq). From three cases of wind speed prediction in Liaoning Province, China, the proposed method was tested from three aspects: accuracy, reliability and training time.

The remainder of this paper is organized as follows. In Section 2, the background of structured knowledge, traditional correlation methods and deep learning methods have been introduced. Then, the optimized-based correlation methods and knowledge-based Seq2Seq architecture have been proposed. In Section 3, compared with traditional deep learning structures, the optimized correlation-based Seq2Seq modeling is applied to the actual wind speed forecasting cases. In Section 4, we summarize the work of this paper has been and give our conclusions.

## 2. Methods and deep knowledge-based learning model

Structured knowledge [38] is the key bridge connecting feature engineering to algorithm selection in the current academic research process. The feature selection of raw data information usually came into notice to statisticians, while in machine learning field, experiments on different algorithms or modeling structures often attract more attention and discussion. Furthermore, some studies commonly increase the complexity of the model rather than analyzing the rationality of data structure or information contained to the model. As a result, the features based on raw data structure cannot match the excessive dimension of models, though some models are probably functional enough. In this paper, we hope to emphasis an underappreciated and undefined section located between the feature engineering and the modeling as shown in Figure 1. Based on the similar studies of this section, we regard it as the knowledge engineering.

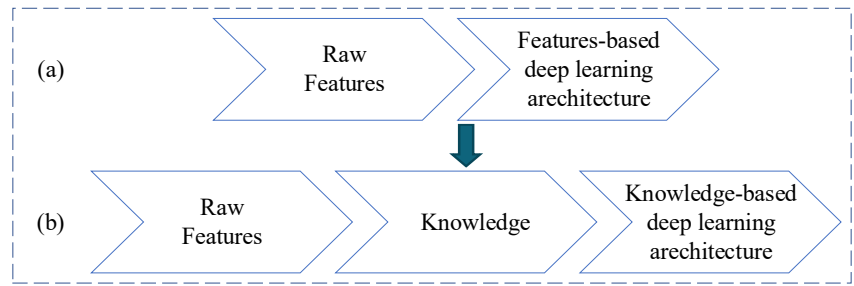

Fig. 1. The architecture of Knowledge-based method.

Intuitively, the feature engineering, knowledge engineering and modeling are attributed to amino acid, polypeptide chain and protein, respectively, corresponding to biological concept. "Features" from feature engineering play central roles on building blocks of models. "Knowledge" looks like a single spatial chain carrying many features, which contains more information than the set of features due to the complex space-time structures. A knowledge-based "Model" comes behind and it will contain one or more "Knowledge".

The above process in deep learning can be regarded as "Perception" [19], which was studied through chess game. The following methodology focuses on the bonds between knowledge engineering and feature engineering and modeling. In order to constitute a chain of features, it is common to introduce correlation analysis methods into the wind prediction model, while the essence of most methods is to consider correlation as a quantitative parameter [20][21] or one of the evaluation methods [6]. With regards to wind data selection problem, Damousis *et al.* noticed its impact on prediction models and they introduced auto-correlation and cross-correlation as the evaluation methods [20]. However, those results are still not satisfactory. In this paper, we moved from a traditional analytical perspective to an optimization perspective to solve this problem so that "knowledge" contains not only the data-structural information but also more advanced information based on statistical optimization.

Similarly in the field of deep learning, the optimal corpus selection is always a puzzling problem need to be solved. Although the state-of-art deep learning methods can reduce the labor work of feature engineering in some domains, it has shown that they are still difficult and restricted to apply steadily in most of industrial problems. The gap between theory and practice comes from the real data and theoretical assumptions. In another word, it could extract the features of variables but it still cannot get the prior knowledge like data correlation. For instance, when it comes to the image recognition field where the deep learning methods show the good performance, it is usually to use convolutional neural network (CNN) and fully connected layer (FCL) [22][23] to take all the corpus or historical information from preprocessed data set as the input data for training model. This operation takes a large amount of time and computing costs, and often requires extremely high hardware support. Encoder-decoder structure is one of the effective solutions for it in industrial problems [24][24]. Therefore, after constructing the knowledge chains of features, a LSTM-based Seq2Seq has been introduced and improved as the suitable architecture to absorb the structural knowledge.

**2.1 Pearson correlation coefficient**

Correlation coefficients are important and widely used statistical methods. Data scientist will consider the strength of the linear relationship between variables by referring to the Pearson correlation coefficient. Particularly, the value of a random variable at each moment is a feature for sequential tasks. Thus, it is helpful to select related features to combine and then fold them into linear chains.

$$\rho_{X,Y} = \frac{\text{cov}(X,Y)}{\sigma_X \sigma_Y} \tag{1}$$

Given a pair of random variables $(X,Y)$, $\rho_{X,Y}$ is Pearson correlation coefficient. $\sigma_X$ and $\sigma_Y$ and the standard deviation of $X$ and $Y$, respectively. $\text{cov}(X,Y)$ represents the covariance between $X$ and $Y$. The variance can also be thought of as the covariance of a random variable with itself.

$$\sigma_X = \sqrt{\text{var}(X)} = \sqrt{\text{cov}(X,X)} \tag{2}$$

Covariance is one of the most important and practical statistical measurements of the joint variability of two random variables [25]. The formula is expressed as below. If the greater values of one variable mainly correspond with the greater values of the other variable, and the same holds for the lesser values, (i.e., the variables tend to show similar behavior), the covariance is positive [26].

$$\text{cov}(X,Y) = E\big[(X - E[X])(Y - E[Y])\big] \tag{3}$$

If the discrete random variable pair $(X,Y)$ can take on the values $(x_i, y_i)$ for $i = 1, \ldots, n$, with equal probabilities $1/n$, then based on Zhang *et al.* 's paper [27], the covariance can be equivalently written as

$$\operatorname{cov}(X,Y) = \frac{1}{n^2}\sum_{i=1}^{n}\sum_{j=1}^{n}\frac{1}{2}(x_i - x_j)(y_i - y_j)$$
$$= \frac{1}{n^2}\sum_{i=1}^{n}\sum_{j>i}^{n}(x_i - x_j)(y_i - y_j). \quad (4)$$

More generally, if there are $n$ different realizations of $(X,Y)$, namely $(x_i, y_i)$ for $i = 1,\ldots,n$, but with different weights $\omega_i$, then the covariance can be written as

$$\operatorname{cov}(X,Y) = \sum_{i=1}^{n}\omega_i(x_i - \bar{x})(y_i - \bar{y}). \quad (5)$$

**2.2 Sequence to Sequence modeling**

Williams *et al.* came up with RNN [28], which forms a very powerful and expressive family for sequential tasks. RNN is not only able to learn the non-linear relationship between historical data and prospective data, but also can effectively improve the prediction accuracy by learning the data variations of adjacent time. All recurrent neural networks have a chain structure of repeating neural network units. The hidden unit at each step in the standard RNN is simply a *tanh* or *ReLU* operation.

Let $x_t$ and $y_t$ be input and output vectors of time series respectively; $h_t$ is the hidden layer vector which represent the state of dynamical system; $W$, $U$ and $b$ are parameter matrices and vector. $f_h$ and $f_y$ are activation functions for hidden and output units. The forward-propagation recurrent network can be described as follows:

$$h_t = f_h(W_h x_t + U_h h_{t-1}) \quad (6)$$

$$y_t = f_y(W_y h_t + b_y) \quad (7)$$

In practical applications, it is common to see that RNNs are difficult to deal with the long term sequence with dependence due to the problem of gradient disappearance or explosion [29]. To address the shortcomings of the standard RNN in long-range contexts modeling, Long Short Term Memory (LSTM) architecture has been put forward by Hochreiter and Schmidhuber [30]. It can be extended effectively in many areas, including weather forecast [31].

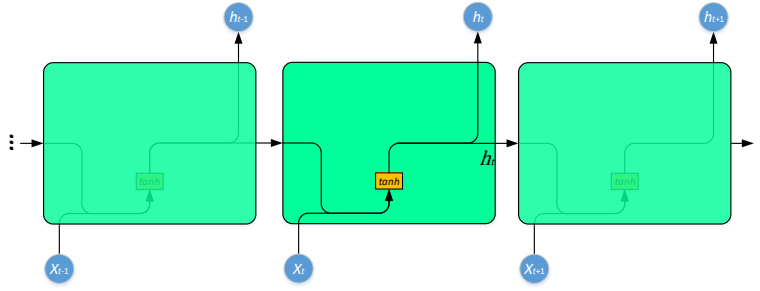

Fig. 2. RNN structure.

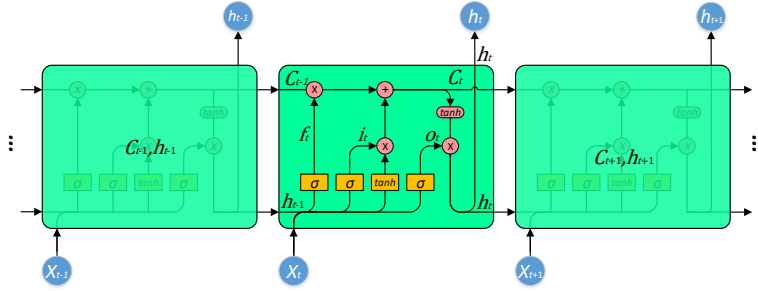

Fig. 3. LSTM structure.

Compared with the structure of RNN, LSTM is augmented by recurrent gates [32]. LSTM contains memory cells to store the temporal state of the network and multiplicative gating units to control the flow of information, which constitutes the basic units in the hidden layer of an LSTM network. There are four neural network layers with three types of gates (forget gate, input gate and output gate) in each memory block of LSTM cycle: three sigmoid layers and one *tanh* layer in Figure 3. During training, forget gate, input gate and output gate, respectively, learns to scale the information of the cell, control the flow of input activations into the memory cell and controls the output flow into the remainder of the network. When gates are closed, irrelevant information and noise do not enter the cell, and the rest of the network will not be perturbed.

$$a_t = \tanh(W_c x_t + U_c h_{t-1} + b_c) \quad (8)$$

$$i_t = \sigma(W_i x_t + U_i h_{t-1} + b_i) \quad (9)$$

$$f_t = \sigma(W_f x_t + U_f h_{t-1} + b_f) \quad (10)$$

$$o_t = \sigma(W_o x_t + U_o h_{t-1} + b_o) \quad (11)$$

$$c_t = i_t \cdot a_t + f_t \cdot c_{t-1} \quad (12)$$

$$h_t = o_t \tanh(c_t) \qquad (13)$$

*W* and *U* are parameter matrices to be estimated during model training; *σ* (*sigmoid*) and *tanh* are activation functions and *b* are vectors for biases. The Eqs (8) - (13) describe the update process of a LSTM cell at each time-step *t*. In the above equations and figure, $x_t$ is the input to memory cell in the current period *t*. $f_t$, $i_t$ and $o_t$ are the values of forget gate, input gate and output gate at time *t*, respectively. $a_t$ is the modulated input to the cell at time *t*. $c_t$ denotes the cell state value at time *t*. $h_t$ is the output of the cell at time *t*.

Sequence to sequence (Seq2Seq) modeling evolved from the recurrent neural networks based encoder-decoder frameworks [33][34]. The initial Seq2Seq mainly focus on the machine translation problems, which turns a national language sequence into another national language sequence. The commonly used recurrent networks in encoder-decoder frameworks are LSTM and gated recurrent units (GRU) [35].

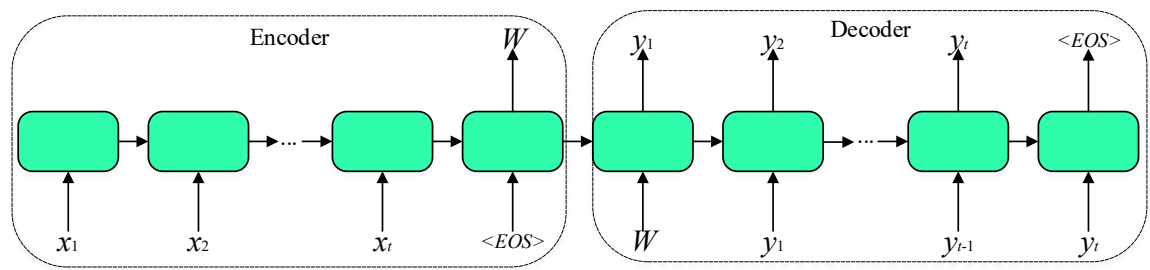

Fig. 4. Seq2seq structure.

In Figure 4, it shows that the sequence $S = (x_1, x_2, ...., x_t)$ has been expanded according to time series for a sequential task. With regard to the general seq2seq structure, it is consist of an encoder-decoder framework, which actually divides a neural network into two parts. In the field of machine translation, the encoder process generally utilizes the neural networks to receive words from the input sequence *S* in time and it will stop receiving the input information until get the target <*EOS*> (End of Sentence). Finally, a vector *W* is output as semantic representation vector of the input sequence.

The single encoder-decoder framework cannot subtly focus on all the targets from input sequences, and therefore cannot perform optimally on the chain structure such as Seq2Seq. Popular choices for optimizing the Seq2Seq models are introducing attention mechanism [36] and bi-directional encoders [34]. Shortcut connections or residual architectures also can improve models with many layers [37].

Although the Seq2Seq modeling can effectively capture the future trend of sequence, it can hardly handle the problem of high precision sequential prediction based on complex features when feature engineering is not introduced. Here, the complex features mean sometimes it cannot obtain direct features that are sufficiently complementary or the prefect mechanism for the predicted target.

**2.3 Knowledge-based Seq2Seq modeling**

Based on the optimized correlation-based model, the knowledge-based methods [38][39] are recognized in this paper, which contributed to artificial intelligent or statistical graphic models. The correlation behaves differently for different issues, either positive or negative, or linear or nonlinear. Correlation coefficient is commonly used to describe the degree of positive and negative linear relationship. As mentioned above, the spatio-temporal correlation analysis methods have been of great value for wind prediction models. Most of research mainly use correlation coefficient as an evaluation index for judging consistency between two different sequences. From authors' knowledge, it is the first time to regard a correlation coefficient as the convex optimized objective from the view of optimization research.

The optimized Pearson correlation model has been proposed to build linear chains of knowledge for Seq2Seq structure based wind speed records from different wind fans in different sites. Then, a novel LSTM-Seq2Seq structure has been designed based on the chain-based knowledge.

**2.3.1 Optimized Pearson correlation model**

Pearson correlation coefficient is not only introduced to describe linear correlation of data but also used as a probabilistic forecasting model. Unlike the probabilistic model based on probability statistics, the optimized Pearson correlation model is object-based modeling through the maximization correlation coefficient as below.

Assume that random variables $X$ and $Y$ are linear correlated. And there is a complete known sequence $X_t$ for variable $X$ at time $t$, while in sequence $Y_t$ for variable $Y$, there exists some unknown data points $Y_t^*$. Then, the correlation model can be built based on Pearson correlation. Thus, according to the Formula (4)-(5), the correlation coefficient $\rho_{X,Y}$ can be converted to objective function $f(Y_t^*)$ by Formula (14).

The optimization problem (14) is a nonlinear fractional programming problem [41].

$$\max_{Y_t^*} \quad f(Y_t^*) = \left| \frac{\frac{1}{n^2} \sum_{i=1}^{n} \omega_i \sum_{j>i}^{n} (x_i^t - x_j^t)(y_i^t - y_j^t)}{\sigma_X \sigma_Y} \right| \quad (14)$$

For energy prediction problems like wind and solar energy, the common fact is that the energy in correlated regions not only shows the same trend, but also has the opposite trend. Therefore, absolute Pearson correlation coefficient has been introduced for wind speed forecasting problems. And then, the historical sequences are completely known and the forecasting sequences can be viewed as the unknown part of the recent sequence. This is obviously a difficult and complex optimization problem due to $\text{cov}(X_t, Y_t)$ and uncertain $\sigma_Y$.

Due to the symmetry of the correlation and the many-to-many input-output structure of Formula (14), it is not only suitable for encoding-decoding sequential structures, but also can be used for univariate time series prediction, multivariate time series prediction, and multi-step time series prediction. Formula (14) can be equivalent to a quadratic fraction programming problem [42].

*Illustrative Case for one-step prediction*

One illustrative solution of Formula (14) is explained with the one-step prediction case. There is a data sample $Y_t$ with missing data $Y_t^*$. With regard to wind speed forecasting problem, the data from time 1 to time $t$ is known, while the last one $y$ of time $t+n$ is unknown. The process in Figure 5 shows that the wind speed of $y_t$ is collected from time 1 to $t+1$ while it needs to forecast the wind speed from $y_{t+1}$ to $y_{t+n}$. And, data sample $X_t$ is completely known, which is high correlated to $Y_t$. In addition, in some practical cases, $x_{t+1}, \ldots, x_{t+n}$ is also unknown but could be predicted within a high trust region. It is also suitable to apply our optimized correlation model. These situations are more complex but commonly exists in extreme weather.

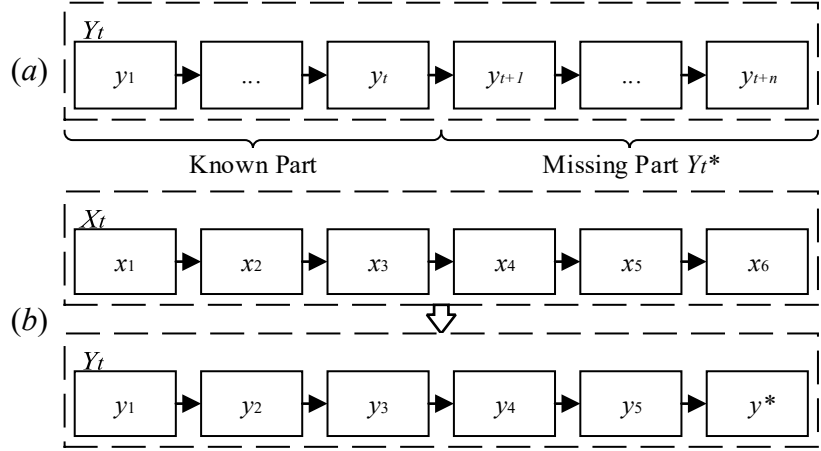

Fig. 5. Illustrative case.

The following step is to predict $y^*$ based on the relationship between data sample $Y_t$ and $X_t$. In this paper, we chose Pearson correlation coefficient as the objective function to describe the linear relationship between data samples. Considering the standard deviation of $Y_t$ need to be estimated first, one of processing methods is to separate vector $Y_t$ into two vectors. The first one is $Y_a = (y_1, y_2, y_3, y_4, y_5, 0)$ and the other is $Y_b = (0, 0, 0, 0, 0, y^*)$. The standard deviation of $Y$ could be given by $Y_a$ and $Y_b$.

$$\sigma_Y = \sqrt{\sigma_{Y_a} + \sigma_{Y_b} + 2\operatorname{cov}(Y_a, Y_b)} \tag{15}$$

And then, the objective function is

$$\max \quad f(y^*) = \left| \frac{\frac{1}{n^2} \sum_{i=1}^{n} \sum_{j>i}^{n} (x_i - x_j)(y_i - y_j)}{\sigma_X \sqrt{\sigma_{Y_a} + \sigma_{Y_b} + 2\operatorname{cov}(Y_a, Y_b)}} \right| \tag{16}$$

With regards to the solution of problem (16), it is such difficult to apply and solve non-convex function in practical applications. Thus, the convex transformation technique has been considered and introduced. By transformation, the mentioned non-convex problem (15) can be transformed equivalently into the following minimized convex problem (16). For one step prediction, the dimension of $y^*$ is one. Thus, it is easy to get the analytic expression of the optimal solution. The detailed derivation for high dimension ($\geqslant$ 2) of $y^*$ is supported in the Appendix.

From this illustrative example, two results can be obtained by the optimized correlation model. According to the input sequence, the most linearly correlated historical sequence can be obtained in statistics. Moreover, there exist different optimal correlated sequences (output sequences) for different stages to one input sequence. After calculating all the Pearson correlation coefficients among input and outputs, it is clear to know the exact relevance degree, which produces an easy white box to analysis the positive and negative relationships among corpus later in the deep learning frameworks as well as the optimization operations such as attention mechanism.

In terms of time-series forecasting of energy, data is potentially associated with sufficient historical data from current database. Although energy forecasting are stochastic problems, a traditional strategy is to refer to the multiple predictions from different methods. Based on the correlation information between the recent and historical data, our proposed method can provide the future predictions on the basis of different historical data conditions. In addition, the structure of the proposed method is synonymous with the probabilistic forecasting methods based on graphical correlation which is called knowledge tree in Section 2.3.2.

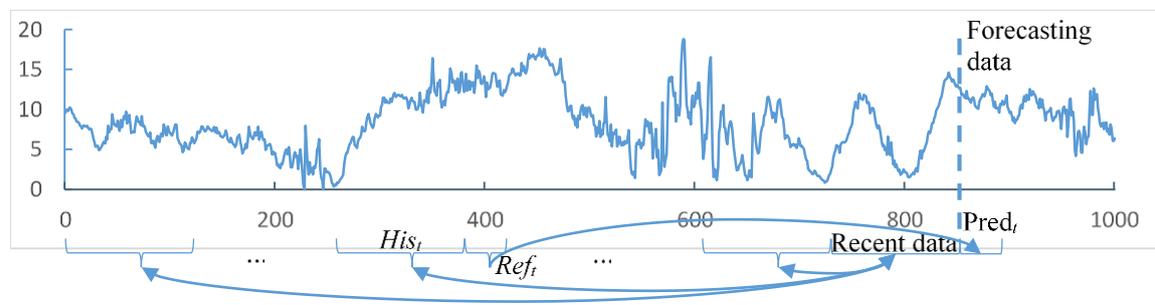

Fig. 6. Illustration for moving window.

Then, based on moving window in Figure 6, the correlated historical sequence $His_t$ for recent data and its referenced sequence $Ref_t$ will be input into the optimized correlation model to give the prediction $Pred_t$. Also, the predictions can be given not only based on the optimized auto-correlation model of target sequence, but also from other related supplementary data sets. The correlation-based predictions with their corresponding sequences are considered as the chains of features for deep learning methods. How to select relevant data and reduce features are discussed in the sequel.

Due to the inevitable contradictions between samples, the feature selection problems are always challenging. On one hand, the model could be underfitting without enough features; on the other hand, if the correlation threshold is defined too low, it is not valuable to give accuracy predictions due to the bad data and time constraints from industries. Thus, it is necessary to decide the optimal width of window and find a threshold to balance the number and the effect of features based on the experience of auto-correlation. As shown in Figure 7 and Figure 8, two correlation coefficients (0.7 and 0.8) are set, resulting in different prediction results.

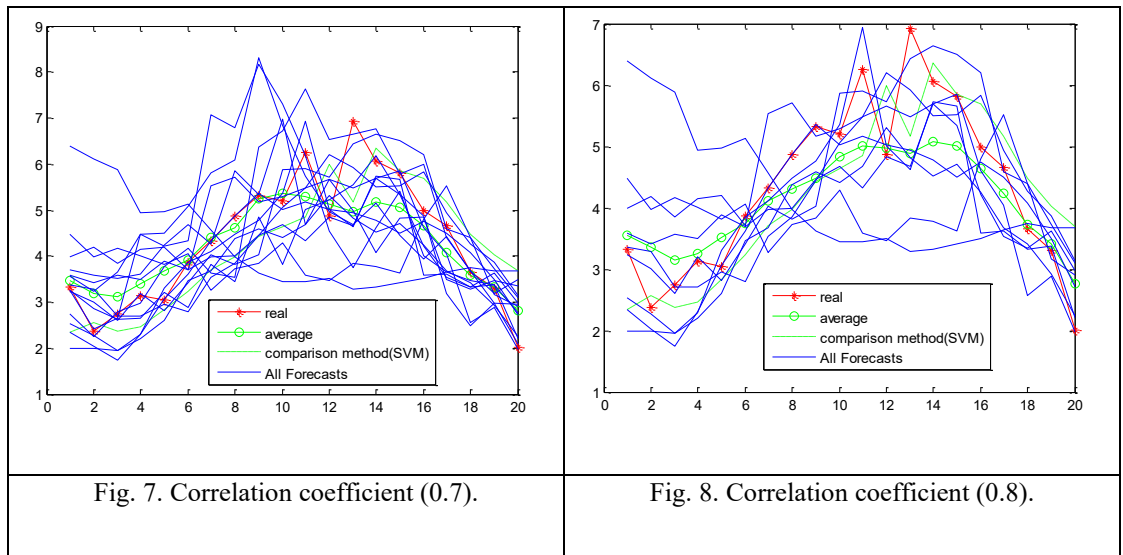

| Fig. 7. Correlation coefficient (0.7). | Fig. 8. Correlation coefficient (0.8). |

According to the results, it can be found that the set of forecasting trends (blue lines) based on different correlated historical data samples can effectively cover range of the future. And by comparing with support vector machine (SVM) method, one of popular machine learning forecasting methods, the correlated-based methods shows a closer trend with the reality.

When it comes to give an ensemble prediction, the existing ensemble models for probabilistic method are generally to use a weighting function to combine all the predictive results with assumption on parallel structure as shown in Figure 11. Both linear (weighted mean method) and nonlinear (entropy-based method) ensemble predictive methods do not perform well in our cases as shown in Figure 9 and Figure 10. One of the key reasonable explanations is that these predictions never stand on the same level. They are more reasonable to be arranged according to the relevance from small to large. The smaller the relevance, the larger the prediction range and the probabilistic forecasting branch is relatively small. The larger correlation coefficient is given, the more specific trend is predicted, and the more likely it is. Thus, as the dialogue problem in natural language processing, the ensemble model for correlated-based predictions should have an architecture similar to that of artificial intelligence conversation system. The specific content will be explained in the following session. Through numeric experiments, it has been found that this structure is not suitable for optimized correlation models. Thus, the structure of sequence to sequence has been introduced based on the similarity with the sematic dialogue systems.

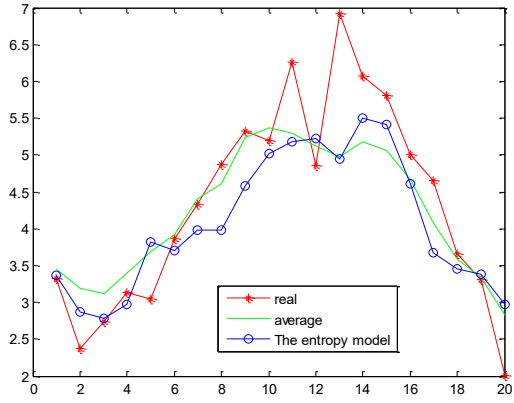 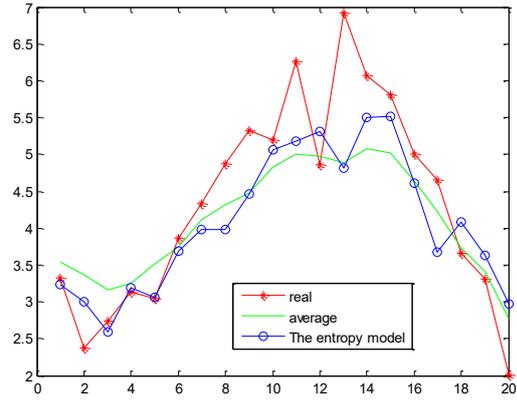

Fig. 9. correlation coefficient (0.7).  Fig. 10. correlation coefficient (0.8).

Considering statistical meaning of the correlation coefficient, it is realized that the predictive results stand on different levels. It is better to choose the structure in the Figure 11(b) instead of that in Figure 11(a). The sequence is ordered by the relationship of $|\rho_1| \leq |\rho_2| \leq |\rho_3| \leq |\rho_4| \leq |\rho_5|$. And it is more structural and powerful to represent logic, if we introduce the multi-layers architecture.

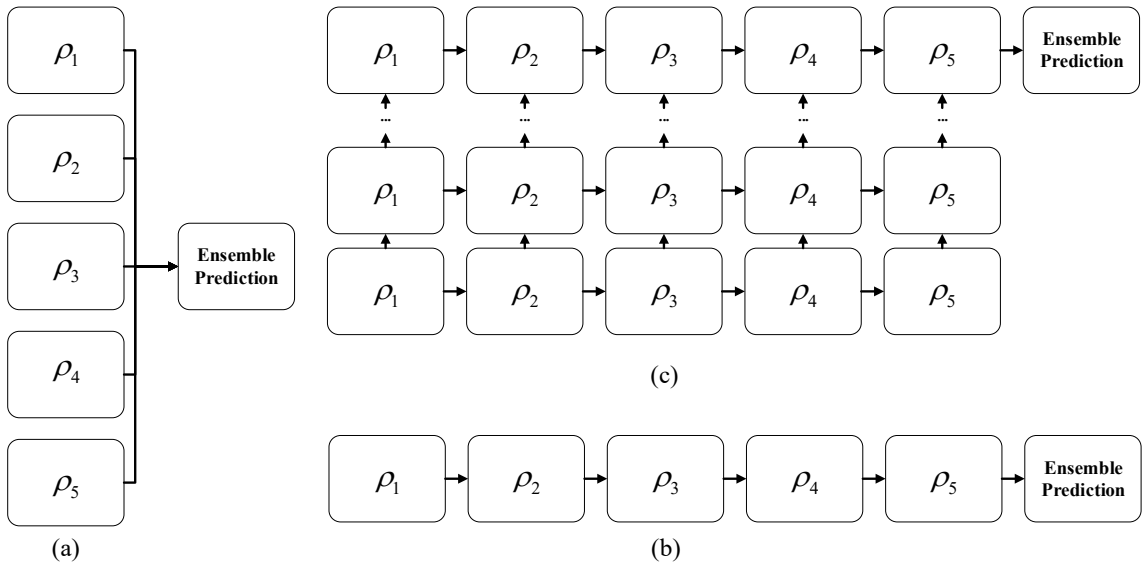

Fig. 11. The structure of the ensemble model for predictions.

In addition, the inference based on the positive and negative linear correlation needs to consider the boundary constraints with precise physical significance.

**2.3.2 Sequence to Sequence modeling based optimized correlated corpus and predictions**

Similarly with the probabilistic forecast methods, each correlated-based prediction can be regarded as one of possible and reasonable results. The probabilistic forecasts concern all the predictions with equal status or stages. Thereby, the weighted mean method is one of the effective ways for it. When it comes to the correlation analysis, each prediction comes from the different views according to different time-space referred data samples. For Seq2Seq modeling, it can be seen as a time series relation from vagueness to concreteness. Taking the above-mentioned into consideration, the deep learning structure has been introduced, which can effectively accommodate the complexity of the wind prediction problems involving the input corpus formed by correlation-based models and bring the satisfactory improvement of the forecasting accuracy.

According to Figure 5, it can be seen that there is structural consistency in the model. If the sentence has been replaced by the data sample of wind speed in a period and meanwhile the question part need to combine more sentences based on the correlation, the structure of the model can be enlarged as shown in Figure 12 where $X_1, \ldots, X_n$ is the sparse correlated samples selected from the whole historical data. $\hat{Y}_{t+\Delta t+n}$ is the sequential predictions $\left(\hat{y}_{t+\Delta t+1},\ldots,\hat{y}_{t+\Delta t+n}\right)$ representing $\nabla t$ time ahead predictions at time $t$. $\hat{Y}_{t+\Delta t}$ is the predictive values before $t+\nabla t$.

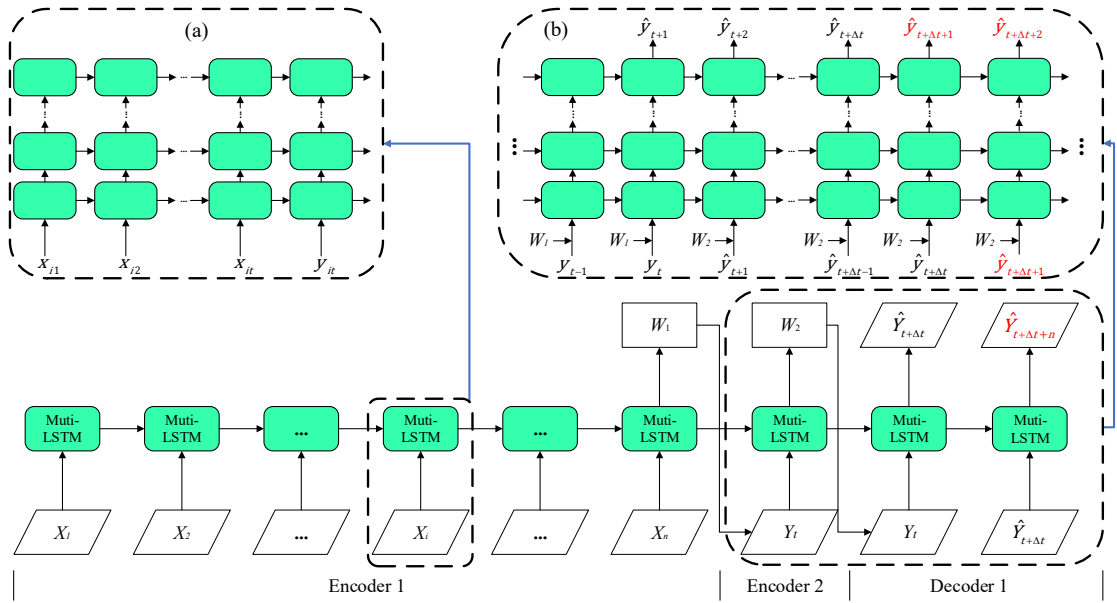

Fig. 12. Structure of optimized correlation-based modeling.

(a) Structure of data sample; (b) Structure of the second encoder and decoder.

The proposed structure of optimized correlation-based modeling includes two part; structure of data sample and structure of the second encoder and decoder. The role of Encoder 1 is to generate several containers for extracted features and build the sparse knowledge tree for sensation region in Figure 13. And Encoder 2 is to record the recent information for predictive objects. In this paper, Encoder 1 is a special structure because each data sample followed by a predictive result $y_{it}$ from optimized correlation

16 / 38

model shown in Figure 12(a). Due to this structure, the proposed model enhances the diversity of predictive functions based on different perspectives. Without $y_{it}$, it is called non-optimized correlation-based structure in this paper. If the Encoder 1 is hidden, the rest structure can be seen as the classical LSTM structure.

Next, from the field of brain cognition, the detailed differences from traditional deep learning are compared. The perceptive process [19] of deep learning structure above is described according to perception of human brain in Figure 13. And Figure 14 shows an illustration with the correlation-based cognition units. The main workflow for human brain is a circle including memory, sensation and cognition, and in the sensation region of brain, it also communicates with the cognition and memory. The designed knowledge tree includes two functions: sensation and cognition. The functions are fulfilled with the relevance model in Section 3.1. Firstly, the sensation of knowledge-based network, as general knowledge-based methods, is extraction and structuring of raw features. Then, the probabilistic predictions with maximizing linear correlation are given referred to the similar historical trends as the cognition function. Finally, the above trends and predictions are reconstructed and memorized as a tree structural knowledge. And knowledge is absorbed by sequence to sequence structure in this paper. Based on CMU, the proposed deep learning framework has a higher dimension than traditional deep learning networks. Each sensation, cognition, and memory unit is made up of traditional deep learning networks (multi-LSTM/GRU) that allow sensation, cognition, and memory units to communicate with each other.

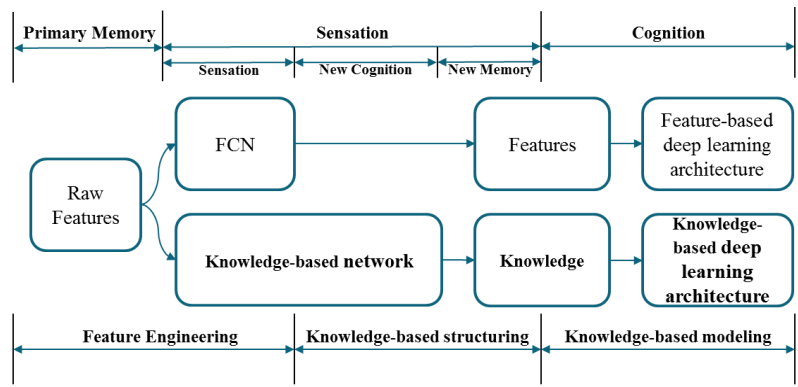

Fig. 13. Comparison between knowledge-based structure and feature-based structure for deep learning.

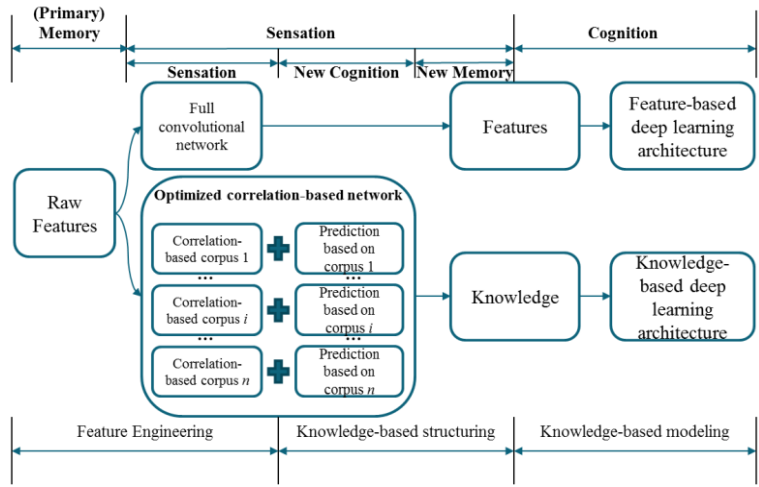

Fig. 14 Knowledge-based structure with correlation-based cognition units.

Specifically, for phenomena of energy transfer and conservation in natural world, such as wind energy, the structure in Figure 16 and 17 can be equivalent to the circular probability network below in Figure 15. Figure 15 and Figure 16 shows the correlation map among all the locations for each time step. Each point on the layer or each color point in the matrix not only denotes the correlation coefficient value but also corresponds to a data sequence. Each circular layer in the structure means the data sequences at one stage from the sensors' record on different locations. And it shows all the data sequences have the probabilistic correlation among all the stages. Usually, such networks contain hundreds of stages and hundreds of nodes on each layer for each stage. Thus, it should be defined as a high dimensional problem.

For instance, a data structure with small size date has been represented in the Figure 15 and 16. And each node of structure in Figure 15 and the point of correlation matrix Figure 16 correspond to 36 stages' wind data from a wind speed segment of a wind farm. The length of each node corresponding to the input data sequence is m, and the length of the predicted data sequence is $n$. Therefore, there are 36 stages and 31 nodes on each layer and the length of each data sequence is ($m+n$), which means the size of the problem would be 36*31*($m+n$) if the FCN methods has been used to introduce the data structure in Fig 14(a) and Fig 15.

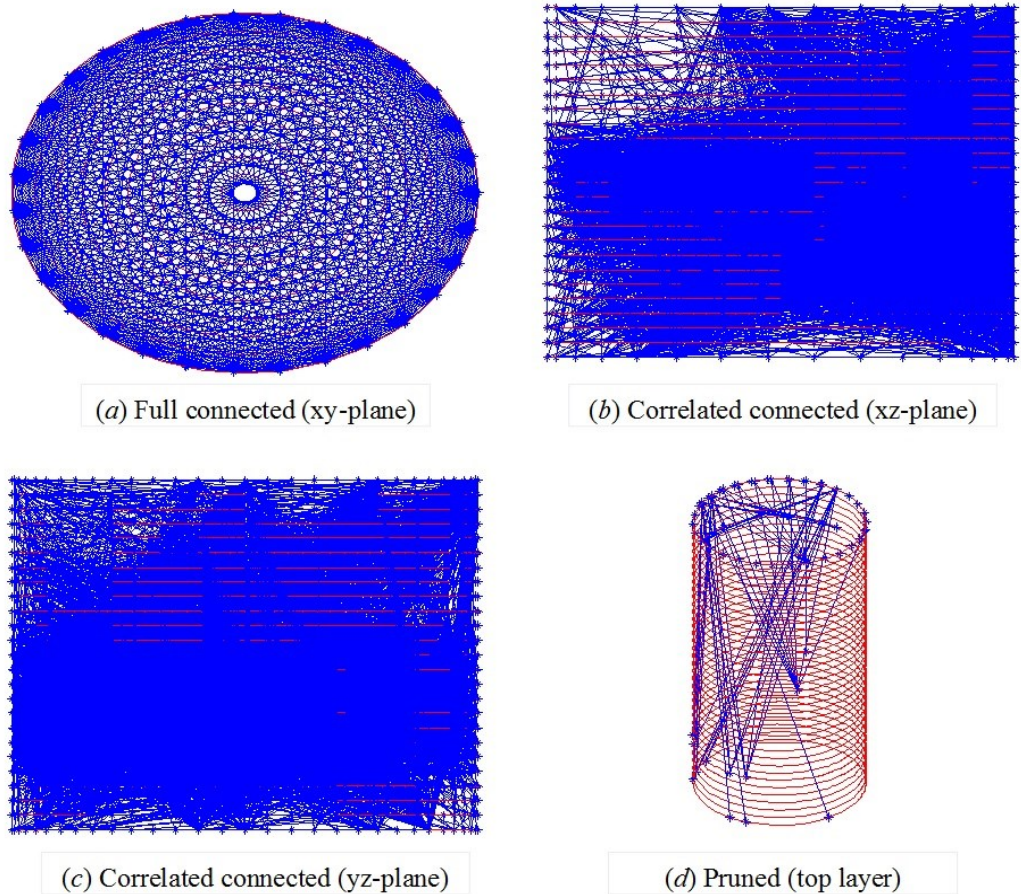

Fig. 15. Knowledge Tree.

In Fig 15, (b) and (c) shows the connections of all points with a correlation above 0.8. The pruned knowledge tree for one layer is shown in (d).

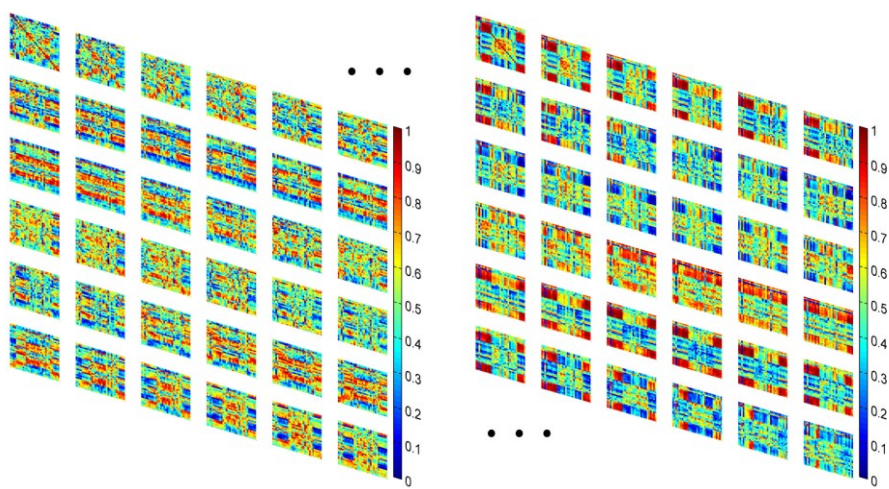

Fig. 16. Knowledge: a probabilistic undirected data network.

In current deep learning studies, researchers often expect the network models to automatically extract data features and then reply on deepening the complexity of the model to train better predictions.

However, the stochastic time-space process like wind or solar always make the deep learning networks overloaded due to the inertia and conservation of energy transfer. Not all the connected points exist in physics or can describe characteristics of objective variables. The correlation reduction of data can not only simplify the input complexity of the model but also reduce the impact of random interference and bad input data on modeling and optimization.

As can be seen from the above Figure 15(d), the structure of knowledge tree can be analogized as a higher-dimensional scenario tree. Thus, the application of pruned knowledge tree not only needs to reduction and reconstruction for modeling, also should consider the text sequence pruning problem [43] [44]. In this paper, a heuristic method for knowledge-based networks has been proposed in Section 3.2.

Different with traditional deep learning methods, the area of Encoder 1 generated by statistical optimization model is not a black box. Thus, after we got the constructed Knowledge Tree, it is easy for researchers to modify it based on some evaluation index, such as calibration and sharpness. Also, by applying the Knowledge-based Seq2Seq model in reverse, a deep learning model based can be built for training the best knowledge tree. In addition, from the perspective of correlation symmetry, it can be found that the structure of knowledge tree is a probabilistic undirected graph model as shown in Figure 17.

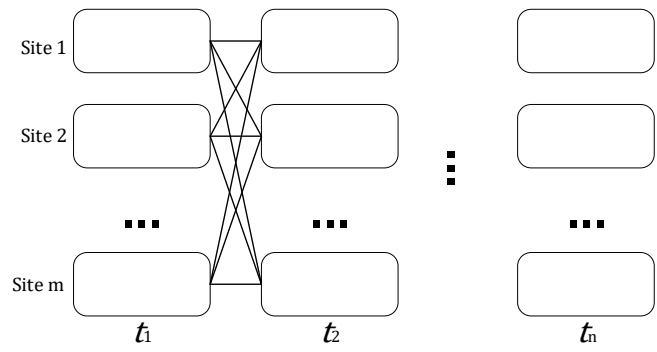

Fig. 17. Probabilistic undirected graph model.

## 3. Experiment

**3.1 Data Introduction**

The data comes from the wind farm in Liaoning Province in China. The Figure 18 is one of the regions of the wind farm where are 33 wind fans and some anemometer towers. In these two figures, the horizontal and vertical coordinates are spaced by 1 kilometer. The $Z$ axis in (b) represents the altitude of the sensors on the fan, and the height difference is distinguished by color. There is a wind speed sensor in each wind fan and wind speed records of all the fans are collected every ten minutes into the database.

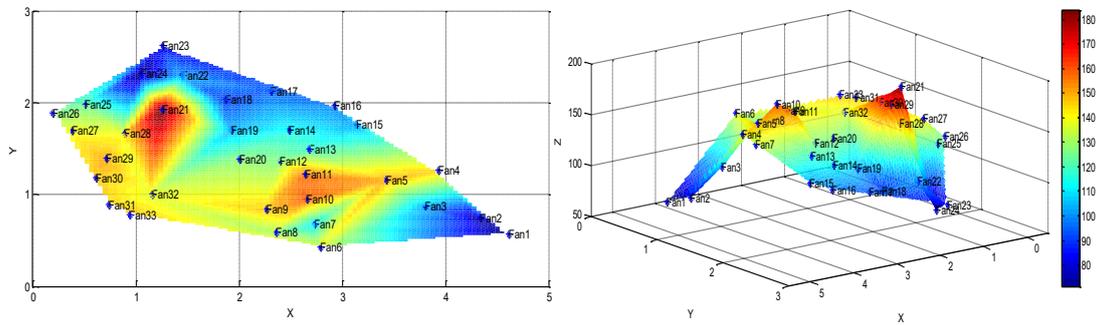

Fig. 18 Geographic map of wind field (a) 2D; (b) 3D

In the numerical experiment, the total length of wind speed data is 31 days from 00:00 on May 29, 2015 to 00:00 on July 1, 2015. Fan 10 is chosen to forecast in this work. Specifically, its location is one of the centers in the wind farm, which means without meteorological information and other sensors, the centrally located fan has more available spatio-temporal reference data.

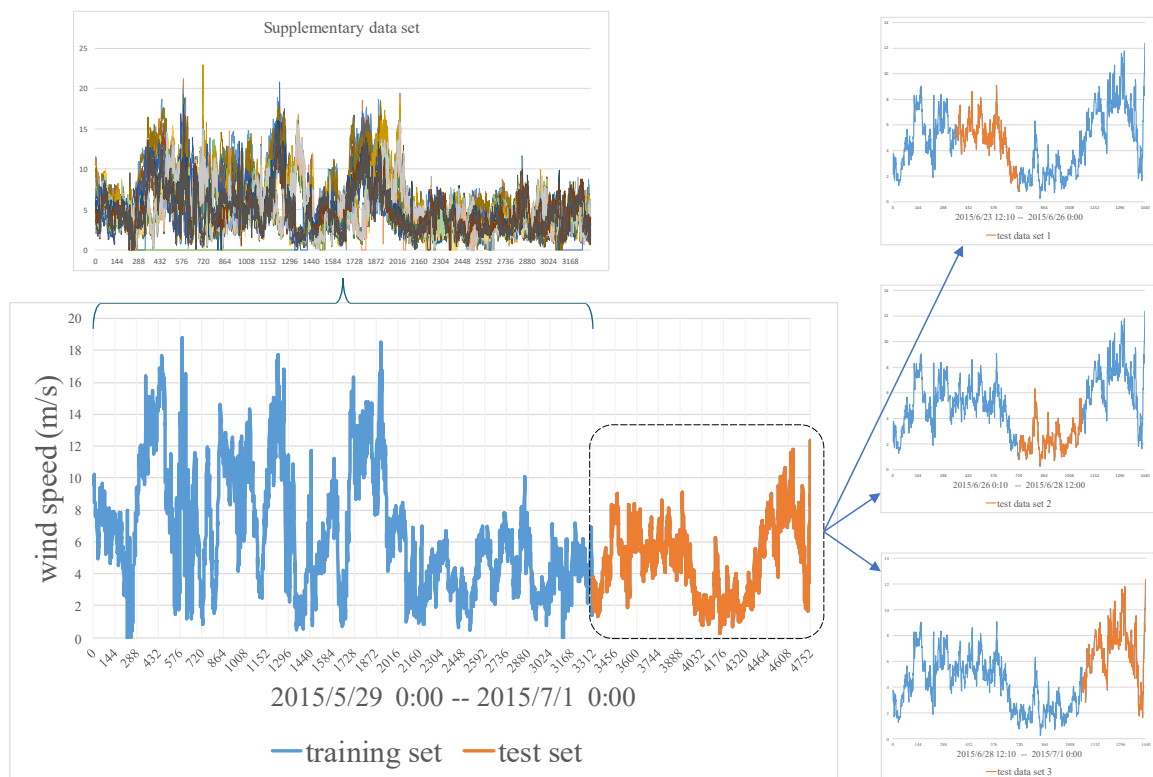

Fig. 19. Training set, test set and supplementary data set.

The historical data of all the sites from 29 May to 20 June have constitutes the supplementary data set, which has been used for optimized correlation-based model to build the knowledge tree for the next 10 days in Encoder 1. Then, three-fold time series cross validation method has been introduced to arrange the 10 days' data into three test datasets as shown in Figure 19. The statistics of three test sets are described in Table 1. *Tr* and *Te* means the length of training set and test set, respectively.

In addition, it can be seen from Figure 19 and Table 2 that the randomness, intermittentness, and instability of the wind speed means that although the training set and test set are adjacent in time, they are commonly different, which is challenging for traditional machine learning. Particularly in Dataset 2 from time series cross validation method, there is an obvious difference between the time series data of the previous period and forecasting trend. And considering that kurtosis and skewness deviate from the normal distribution and the low 25% quartile (Q1) of wind speed, it is always difficult to predict this fluctuating downward trend accurately without sufficient training data or constraints.

**Table 1.** Statistical description of three test datasets

| Dataset unit | Time 1 period=10 min | $T$ period | $Tr$ | $Te$ | min m/s | mean m/s | max m/s | std |
|---|---|---|---|---|---|---|---|---|
| **Dataset 1** | 2015/6/21 0:00 -- 2015/6/25 23:50 | 720 | 360 | 360 | 0.80 | 5.0251 | 9.07 | 1.7048 |
| **Dataset 2** | 2015/6/21 0:00 -- 2015/6/28 11:50 | 1080 | 720 | 360 | 0.25 | 4.1396 | 9.07 | 1.9678 |
| **Dataset 3** | 2015/6/21 0:00 -- 2015/6/30 23:50 | 1440 | 1080 | 360 | 0.25 | 4.7973 | 11.79 | 2.2858 |
| **Supplementary dataset** | 2015/5/29 0:00 -- 2015/6/20 23:50 | 3312 | -- | -- | 0.00 | 6.6252 | 18.80 | 3.86219 |

**Table 2.** Distribution statistics of three test datasets

| Dataset | Time | Skewness | Kurtosis | Q1 | Q3 |
|---|---|---|---|---|---|
| **Dataset 1** | 2015/6/21 0:00 -- 2015/6/25 23:50 | -0.54 | 0.07 | 4.28 | 5.95 |
| **Dataset 2** | 2015/6/21 0:00 -- 2015/6/28 11:50 | 0.89 | 0.71 | 1.67 | 2.94 |
| **Dataset 3** | 2015/6/21 0:00 -- 2015/6/30 23:50 | -0.19 | -0.04 | 5.31 | 7.96 |

**3.2 Generating knowledge tree**

For the wind prediction problem, the network in Figure 15(a) is the original structure of collected data, which can be regarded as a higher-dimensional scenario tree [11]. As a result, the computing ability is always limited due to the high dimension. In practical operations, it is impossible to use FCN method to load the whole networks into deep learning models. Therefore, a heuristic algorithm is proposed to construct "knowledge" from the architecture in Figure 15 to that in Figure 16.

Step1: Set a default threshold for the correlation coefficient $X$, the maximum neighbor nodes $Y$ in one layer and the chain length of encoder in Seq2Seq model.

Step2: Calculate the correlation coefficients between each data sample and target data sample in each layer. If the number of the neighbor nodes is more than $Y$ in $i$ th layer, then the top $Y$ correlated nodes of this layer will be selected and these nodes will be preserved and chosen as the input for Pearson correlation-based model; otherwise, this layer will be ignored.

Step3: Delete the similar layers based on their similarity.

Step4: Check the quantity of selected data samples in different layers. If it is more than $Z$, choose top $Z$ original sequences first and input them into Pearson correlation-based model, respectively. The input corpus for Seq2Seq model consists of the top $Z$ original sequences and results from optimized correlation model. Then, procedure will be terminated; otherwise, relax the default thresholds and go to Step 2.

## 3.3 Model selection

The deep learning methods mentioned above and our proposed methods have been compared through Dataset 1, 2 and 3. The prediction accuracy of optimized correlation-based Seq2Seq method has been verified by the evaluation of 1 hour ahead point forecasting result evaluation. The root mean square error (RMSE) function 错误!未找到引用源。 with L2-norm has been used as loss function for the point prediction of LSTM model and Seq2Seq model described in Formula (17) and (18), respectively. For sequence to sequence structure, $RMSE(\omega; X, Y_{t+\Delta t})$ is the loss function for the time between current time and predicted time.

$$L(\omega; X, Y_{t+\Delta t}) = RMSE(\omega; X, Y_{t+\Delta t}) + \lambda \|\omega\|_2^2 \quad (17)$$

$$L(\omega; X, Y_{t+\Delta t}, Y_{t+\Delta t+n}) = \alpha_1 RMSE(\omega; X, Y_{t+\Delta t+n}) + \alpha_2 RMSE(\omega; X, Y_{t+\Delta t}) + \lambda \|\omega\|_2^2 \quad (18)$$

As shown in Figure 12, $Y_{t+\Delta t}$ is the historical wind speed sequence $(y_{t+1},...,y_{t+\Delta t})$ and $Y_{t+\Delta t+n}$ is the historical wind speed sequence $(y_{t+\Delta t+1},...,y_{t+\Delta t+n})$. $X$ is the structural training samples based on knowledge tree.

Point prediction metrics of six models in three datasets are shown in Table 3. Generally, LSTM and GRU have very similar prediction performance for time series. However, in this experiment, the optimized correlation-based model (GRU-based) performed better on dataset 2. Therefore, the experiment mainly compares the optimized correlation-based Seq2Seq method (GRU-based) with non-optimized correlation-based Seq2Seq method (GRU-based) and LSTM method. The best testing results are highlighted with yellow fill. For prediction accuracy and R squared, optimized correlation-based Seq2Seq (GRU-based) has the best results in Dataset 1 and Dataset2. For RMSE, optimized correlation-based Seq2Seq (GRU-based) gets the best results in three datasets.

## 3.4 Evaluation metric

Accuracy (ACC), RMSE and R squared (R2) value have been recorded as the evaluation scores for optimized correlation-based Seq2Seq (GRU-based), LSTM, non-optimized correlation-based Seq2Seq (GRU-based), optimized correlation-based Seq2Seq (LSTM-based), LSTM-based Seq2Seq, and GRU-based Seq2Seq methods. The evaluation scores of ACC, RMSE and R2 have been shown in formula (19)-(21).

$$ACC=\left|\frac{\hat{y}_i - y_i}{y_i}\right|\times 100\% \tag{19}$$

$$R2=1-\frac{\sum_{i=1}^{Te}(\hat{y}_i - y_i)^2}{\sum_{i=1}^{Te}(\bar{y} - y_i)^2} \tag{20}$$

$$RMSE=\sqrt{\frac{1}{Te}\sum_{i=1}^{Te}(\hat{y}_i - y_i)^2} \tag{21}$$

$y_i$ is the actual wind speed in the test set, $\hat{y}_i$ is the prediction and $\bar{y}$ is the average of the real values.

### 3.5 Comparison

The prediction results are compared by the evaluation scores, training time and convergence. The following Figure 18 shows the average loss function, the red curve is the average loss function of training data, and the blue curve is the average loss function of test data.

For the first three methods, their best, worst, and median scores have been recorded in Table 3 from eleven experiments during training and testing, respectively. And the last three methods show median scores only.

For prediction accuracy, the worst case of optimized correlation-based Seq2Seq (GRU-based) in dataset 2 is 85.88% on average. However, the median accuracy of the worst case is 92.30%. LSTM perform the worst in three cases, especially in Dataset 2. The main reason is the different fluctuate trend. For ordinary time series sequential forecasting problem, LSTM usually need to expand the training set and enlarge the window to solve this situation. However, it is not feasible in wind speed forecasting due to the seasonality and auto-regression problems [40]. The LSTM is exposed to one input at a time with no fixed set of lag variables, as the windowed-multilayer Perceptron (MLP). On one hand, it hard to know the perfect training set in advance and if the window size is bigger, the average accuracy of deep learning networks will decrease because it introduces more noise and bad data.

For these three datasets, all the correlation-based Seq2Seq methods finished training within 90s for 500 epochs and 60s for 200 epochs with Intel Core i5-5200U 2.20 GHz CPU and 16.0 GB 1600 MHz RAM. To the methods without the knowledge tree structure, they finished training within 60s for 500 epochs and 30s for 200 epochs. If the methods without the knowledge tree structure add more recent data from supplementary set to the training set, the training time will multiply according to the size of training data without the accuracy improvement.

**Table 3. Results of point prediction**

1 hour ahead prediction

| | | | optimized correlation-based Seq2Seq (GRU-based) | | | LSTM | | | non-optimized correlation-based Seq2Seq (GRU-based) | | |
|---|---|---|---|---|---|---|---|---|---|---|---|
| | | | ACC | RMSE | R2 | ACC | RMSE | R2 | ACC | RMSE | R2 |
| | | max | 98.17% | 0.0988 | 0.9971 | 98.43% | 0.0896 | 0.9980 | 96.46% | 0.1877 | 0.9894 |
| | train1 | med | 96.72% | 0.1774 | 0.9905 | 97.82% | 0.0982 | 0.9967 | 98.43% | 0.0859 | 0.9978 |
| Data | | min | 95.58% | 0.2551 | 0.9752 | 97.71% | 0.1581 | 0.9917 | 95.31% | 0.2124 | 0.9864 |
| set 1 | | max | **97.16%** | **0.1608** | **0.9896** | 88.83% | 0.6076 | 0.8521 | 96.22% | 0.1993 | 0.9840 |
| | test1 | med | 96.13% | 0.2014 | 0.9837 | 79.11% | 0.9794 | 0.6156 | **95.96%** | **0.1854** | **0.9862** |
| | | min | 95.74% | **0.2088** | **0.9824** | 74.89% | 1.0761 | 0.5359 | **95.93%** | 0.2370 | 0.9774 |
| | | max | 96.00% | 0.2126 | 0.9879 | 98.26% | 0.1044 | 0.9965 | 96.95% | 0.1488 | 0.9920 |
| | train2 | med | 96.35% | 0.2195 | 0.9932 | 97.98% | 0.1160 | 0.9959 | 96.14% | 0.1868 | 0.9874 |
| Data | | min | 95.18% | 0.2010 | 0.9862 | 97.23% | 0.1629 | 0.9907 | 97.98% | 0.0994 | 0.0994 |
| set 2 | | max | **89.96%** | **0.1658** | **0.9679** | 37.04% | 0.9315 | 0.2111 | 88.60% | 0.2102 | 0.9597 |
| | test2 | med | 87.06% | 0.2358 | 0.9493 | 10.35% | 1.0376 | 0.0210 | **88.52%** | **0.2122** | **0.9590** |
| | | min | 85.88% | **0.1498** | **0.9762** | -8.07% | 0.9398 | 0.1970 | **86.11%** | 0.2514 | 0.9424 |
| | | max | 96.54% | 0.1396 | 0.9956 | 97.27% | 0.1092 | 0.9962 | 98.37% | 0.0703 | 0.9987 |
| | train3 | med | 95.89% | 0.1923 | 0.9876 | 97.03% | 0.0884 | 0.9983 | 97.93% | 0.0848 | 0.9981 |
| Data | | min | 97.76% | 0.1105 | 0.9965 | 96.61% | 0.1279 | 0.9964 | 96.92% | 0.9692 | 0.9938 |
| set 3 | | max | 96.39% | **0.3582** | 0.9400 | 79.13% | 1.5887 | 0.3860 | **96.78%** | 0.4551 | **0.9493** |
| | test3 | med | 96.20% | **0.4124** | 0.9420 | 75.52% | 1.7240 | 0.2769 | **96.43%** | 0.4684 | **0.9463** |
| | | min | 95.60% | **0.5105** | 0.9242 | 54.82% | 0.0082 | 2.0191 | **95.97%** | 0.5441 | **0.9276** |

|  |  | optimized correlation-based Seq2Seq (LSTM-based) | | | LSTM-based Seq2Seq | | | GRU-based Seq2Seq | | |
|---|---|---|---|---|---|---|---|---|---|---|
| Data set 1 | med test 1 | 96.98% | 0.1489 | 0.9911 | 80.75% | 1.0469 | 0.5583 | 80.28% | 1.0483 | 0.5571 |
| Data set 2 | med test 2 | 75.91% | 0.4390 | 0.8243 | 63.89% | 0.9631 | 0.1542 | 63.81% | 0.9822 | 0.1203 |
| Data set 3 | med test 3 | 96.52% | 0.4820 | 0.9432 | 78.89% | 1.7594 | 0.2427 | 78.74% | 1.7831 | 0.2222 |

The convergence curves of optimized correlation-based method in Dataset 1 are shown in the Fig. 20. Adam, Root Mean Square Prop (RMSprop) and gradient descent optimization algorithms are used to verify the convergence. According to training R squared and RMSE curves, it shows a stable feature when training with the loss function in Formula (20). The average loss function and the average prediction accuracy function show a trend of decline and increase respectively with the increase of the number of learning epochs and also the accuracy curves of the training and testing accuracy shows good fitting properties of the proposed methods.

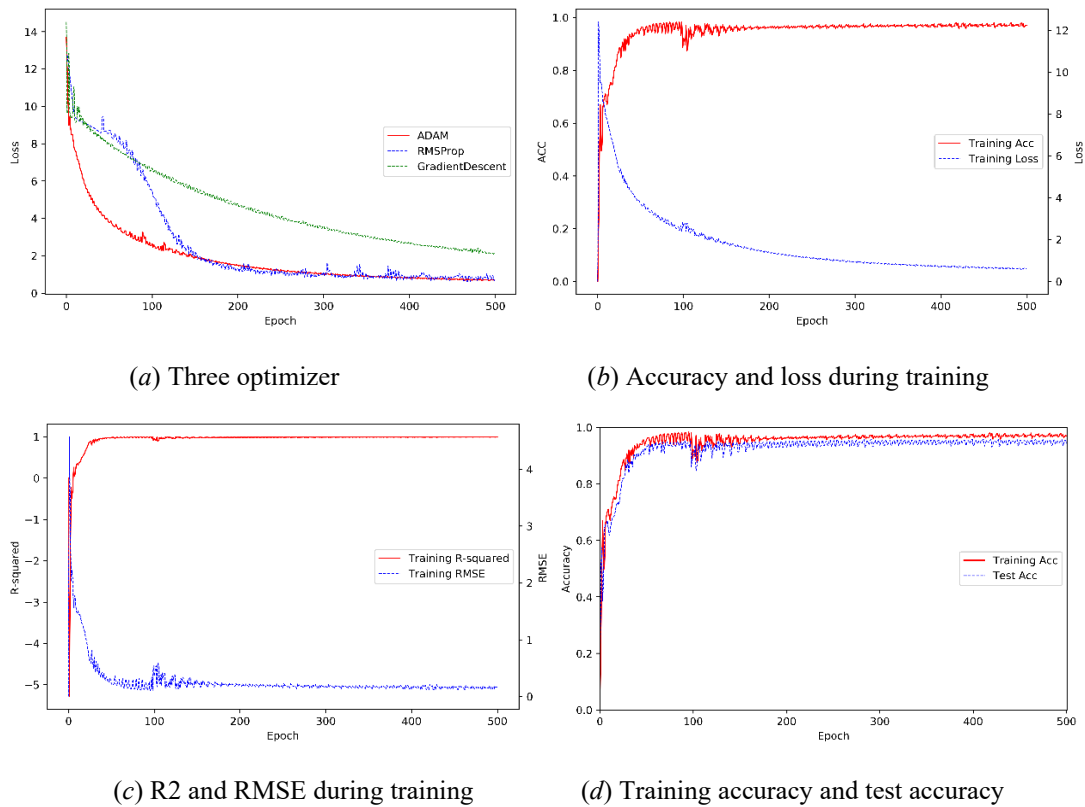

(a) Three optimizer     (b) Accuracy and loss during training

(c) R2 and RMSE during training     (d) Training accuracy and test accuracy

Fig. 20. Convergence curve

In order to compare the differences in point predictions more visually, the prediction results of six models in three test sets are drawn in the Figure 21. Optimized correlation-based Seq2Seq (GRU-based), LSTM, non-optimized correlation-based Seq2Seq (GRU-based), optimized correlation-based Seq2Seq (LSTM-based), LSTM-based Seq2Seq, and GRU-based Seq2Seq methods use the median results as the final prediction result.

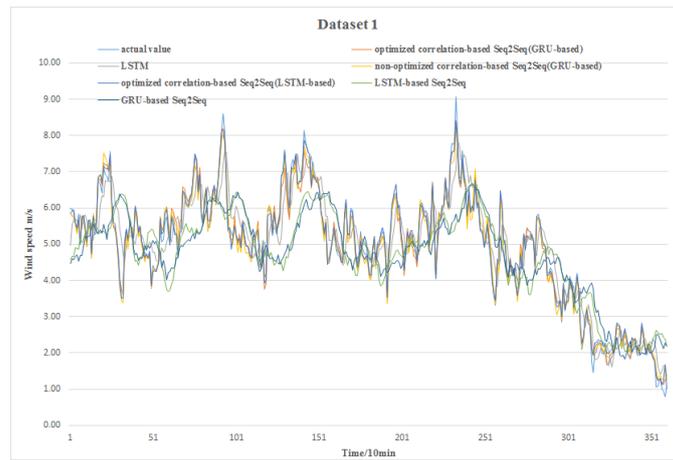

(a) Prediction results of six models in dataset 1

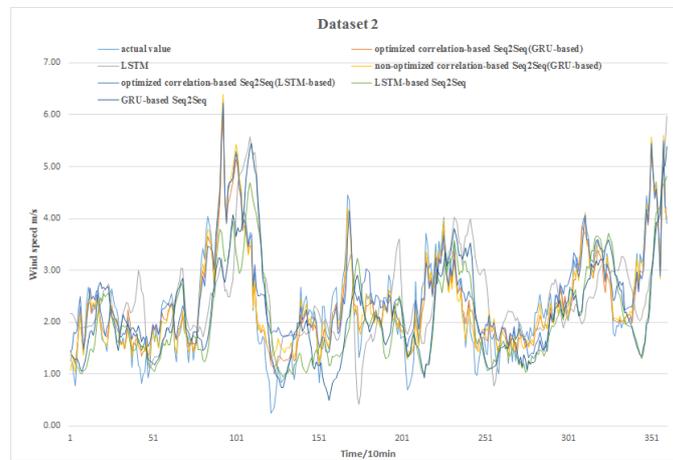

(b) Prediction results of six models in dataset 2

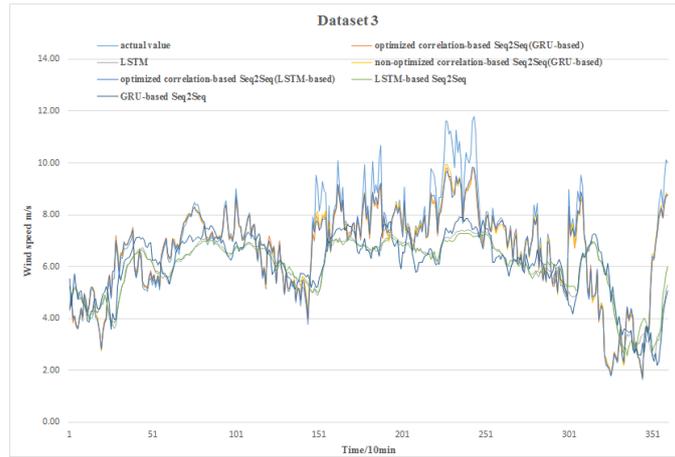

(c) Prediction results of six models in dataset 3

Fig. 21. 1 hour ahead prediction results of six models in three test sets

Compared with the proposed method, it shows the unstable performance from the median LSTM case in Dataset 2 as shown in Figure 22. According to the limited input information, traditional machine learning methods such as LSTM method are hardly to judge the difference between future and history and likely to give a myopic prediction.

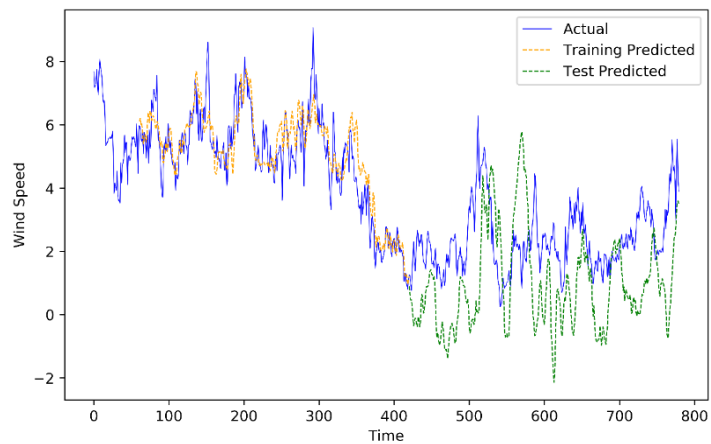

Fig. 22 Training and test prediction of LSTM

### 3.6 Discussion

*1) Non-optimized correlation-based model*

The correlated corpus selection based on non-optimized correlation models is essentially a dimensionality reduction of input data. When the historical data and the scale of knowledge are sufficient enough, it is equivalent to the optimal feature extraction with FCN. The advantages are described as follows.

1. Each output corresponds to the degree of visible input corpus and correlation. In the development of the input-output relationship, from the black box to the white box, it is convenient to maintain and correct the deep learning model, and to ensure its reliability requirements in the industrial field.
2. Due to the visible encoder structure for knowledge tree, the input part also can be an output to be trained as the decoder. Thus, it is creative to build two deep learning networks to supervise each other so that it can find and establish a better function to describe the causal relationship based on data from cause to effect and from effect to cause.
3. Greatly reducing the data dimension, deep learning can be trained with CPU. Time is saved, and on the other hand, accuracy is guaranteed. According to these two aspects, the corpora based on optimized correlation model make the Seq2Seq structure more suitable for energy prediction problems.
4. Solving the lack of feature engineering methods in general sequence to sequence models

*2) Optimized correlation-based model*

Compared with Non-optimized correlation-based model, the encoder for knowledge tree contains different knowledge-based predictive information based on the correlations between historical data and recent data.

It not only inherits the above advantages of non-optimized correlation-based models but also the similar functions as the probabilistic forecasting methods.

Compared with the traditional deep learning structure, the proposed structure has one or more knowledge layers. Optimized correlation-based model is outstanding due to the layer of "knowledge tree". For the Knowledge layer, the optimized correlation-based model we proposed is a method based on prediction objects. Specifically explained, a single knowledge point (single peptide chain) is formed by optimally predicting the target from the perspective of multiple linearly related historical objects. Then, through comprehensive consideration (folding), knowledge (polypeptide chain) is formed. Traditional deep learning does not have such a personalized prediction mode. It can only use a lot of hardware and invest more time in training. It is expected to obtain a model that is super complicated enough to deal with all future situations. This expectation is not consistent with actual engineering applications. Therefore, our method performs more prominently in data-based feature extraction and dimensionality reduction and data-based modeling.

In addition, the optimized correlation-based method has some other advantages according to our experience.

1. According to the introduction of features, it is a good method to establish the formation of knowledge based on the optimized correlation-based model.

2. It can be found that for ultra-short-term predictions (15 minutes in advance), the structure of the six methods is similar and the prediction results are almost the same. With the increase of the prediction interval, the proposed method is more stable than the LSTM and seq2seq methods. It not only grasps the trend correctly, but also predicts the "inflection point". The traditional time series problem always has a lag at the inflection point.

3. For the wind speed prediction problem, the accuracy of LSTM depends heavily on the training set. The length of the training set is seriously affected. If the length is long, training will be slow, introduce noise, and be inaccurate. If the length is short, the upper and lower amplitudes are susceptible to the shape of the training set. Optimized correlation-based model is equivalent to introducing a sparse training set to solve this problem

## 4. Conclusion and future work

Knowledge-based networks can effectively improve the traditional structure of artificial intelligence. In this paper, a convex optimized correlation-based method is proposed for the structural knowledge from optimization view. Compared with the traditional deep learning networks, the proposed deep learning framework introduced the sparse correlation-based cognition units and memory units which consist of several deep networks. Thus, units can Three real cases are given to verify that the knowledge tree established by optimized correlation-based model make Seq2Seq network more stable, accurate and efficient in the stochastic process such as wind speed prediction.

Future work:

1. The structural knowledge-based methodology has the potential for spatiotemporal prediction, such as univariate time series prediction, multivariate time series prediction, and multi-step time series prediction.

2. It is worthwhile to develop other statistical methods from an optimization perspective. Also, the construction of the knowledge tree will be improved and optimized.

3. The proposed method can not only be used to predict wind speed, it is suitable to expand the knowledge-based spatial-temporal correlation models to other spatial-temporal topics, such as disease prevention and control.

## Acknowledge

# Appendix

The general solution for non-linear fraction programming problem is given in paper [42].

1) For one dimensional quadratic fraction programming problem to one-step time series forecasting based on optimized correlation function, the problem is expressed as below.

$$f(x) = \frac{ax^2 + bx + c}{dx^2 + ex + f}(ab \neq 0)$$

$$f'(x) = \frac{(ae-bd)x^2 + 2(af-cd)x - ce + bf}{(dx^2 + ex + f)^2}$$

When the function is extreme, the derivative is 0.

$$(ae-bd)x^2 + 2(af-cd)x - ce + bf = 0$$

The expression of the optimal value can be obtained by simultaneous equations.

$$\begin{cases} y = \frac{ax^2 + bx + c}{dx^2 + ex + f}(ab \neq 0) \\ (ae-bd)x^2 + 2(af-cd)x - ce + bf = 0 \end{cases}$$

$$(e^2 - 4df)y^2 + (4(af-cd) - 2be)y + b^2 + 4ac = 0$$

Then, by comparing the extreme value of the correlation coeffiencient, the maximum value and the optimal solution can be obtained.

2) For high dimensional problem, bisection method can be used for quasiconvex optimization problem as below.

$$f_0(x) = \frac{p(x)}{q(x)}$$

with $p$ convex, $q$ concave, and $p(x) \geq 0$, $q(x) > 0$ on **dom** $f_0$ can take

$$\phi_t(x) = p(x) - tq(x):$$

- for $t \geq 0$, $\phi_t$ convex in $x$

- $\dfrac{p(x)}{q(x)} \leq t$ if and only if $\phi_t(x) \leq 0$

- for fixed $t$, a convex feasibility problem in $x$

- if feasible, we can conclude that $t \geq p^*$; if infeasible, $t \leq p^*$

**Quasiconvex optimization via convex feasibility problems**

$$\begin{aligned} \max \quad & \phi_t(x) \leq 0 \\ \text{s.t.} \quad & f_i(x) \leq 0 \quad i=1,\mathrm{K},m \\ & Ax = b \end{aligned} \quad (19)$$

*Bisection method for quasiconvex optimization*

**given** $l \leq p^*$, $u \geq p^*$, tolerance $\varepsilon > 0$.

**repeat**

1. $t := (l+u)/2$.

2. Solve the convex feasibility problem (19).

3. **if** problem (19) is feasible, $u := t$; **else** $l := t$.

**until** $u - l \leq \varepsilon$

When the sub-problem is non-convex, the relaxation technique method [42] can be introduced or the straightforward algorithm [45] can be used.

For wind prediction, the lower and upper bound is between 0 and 1, and can be set by priori knowledge from correlation analysis. The convex feasibility sub-problem (19) in wind prediction is semidefinite quadratic linear programming problem [46].